\pdfoutput=1
\documentclass[runningheads]{llncs}

\usepackage{eccv}

\usepackage{eccvabbrv}

\usepackage{graphicx}
\usepackage{booktabs}
\usepackage{siunitx}

\usepackage[accsupp]{axessibility}  

\usepackage{hyperref}

\usepackage{orcidlink}

\begin{document}

\title{Follow Your Track: Precise Skeleton Animation Controlled by 3D Trajectories } 

\titlerunning{Follow Your Track}

\author{Yueting Liu\inst{1,2}\orcidlink{0009-0009-3712-3227} 
\and
Yanqin Jiang\inst{2,3}\orcidlink{0009-0003-7338-7963} 
\and
Nian Liu\inst{2,3}\orcidlink{0009-0004-2342-6749}
\and
Jingmen Zhou\inst{2,3}\orcidlink{0000-0003-2936-0811}
\and 
Zhengjun Zha\inst{1}\orcidlink{0000-0003-2510-8993}
\and 
Weiming Hu\inst{2,3}\orcidlink{0000-0001-9237-8825}
\and 
Jin Gao\inst{2,3}\orcidlink{0000-0002-8925-5215}
}

\authorrunning{Y. Liu et al.}


\institute{School of Information Science and Technology, University of Science and Technology of China, Anhui 230026, China\and
State Key Laboratory of Multimodal Artificial Intelligence Systems (MAIS), CASIA, Beijing 100190, China \and
School of Artificial Intelligence, University of Chinese Academy of Sciences, Beijing 100190, China
}

\maketitle

\begin{abstract}
4D generation aims to animate 3D objects with realistic motion, holding great promise for applications. Existing methods typically decouple 3D asset generation from motion synthesis: acquire a 3D asset, prepare a structural representation like mesh and Gaussians, and synthesize motion from text or video control signals.
However, dense mesh and Gaussian representations incur high computational costs and are prone to temporal artifacts, limiting animation quality and duration to only short clips. Meanwhile, text lacks fine-grained spatial and temporal details such as timing and coordination, while video entangles motion with appearance and background. Together, these limitations result in 4D animations that suffer from poor temporal consistency, wrong identification, and limited controllability.
We address these issues with \texttt{ACT}, a trajectory-conditioned framework for topology-general skeletal animation. ACT uses skeletons as a compact structured and compute-efficient representation and 3D point trajectories from monocular video as explicit motion guidance  which provide detailed motion patterns without appearance entanglement. At the core of ACT is a Routed Trajectory Injector, which achieves accurate and robust trajectory-to-joint transfer through three complementary designs: prior-guided hard routing establishes precise skeleton-to-mesh correspondences, global routing enables holistic joint-track interaction for full-body motion awareness, and local windowed cross-attention enforces fine-grained temporal alignment, improving micro-timing and reducing motion misalignment across varying motion rates.
Extensive experiments demonstrate that \texttt{ACT} significantly outperforms existing methods in fidelity and temporal consistency.
\end{abstract}

\section{Introduction}

\begin{figure}[t] 
  \centering
  \includegraphics[width=\columnwidth]{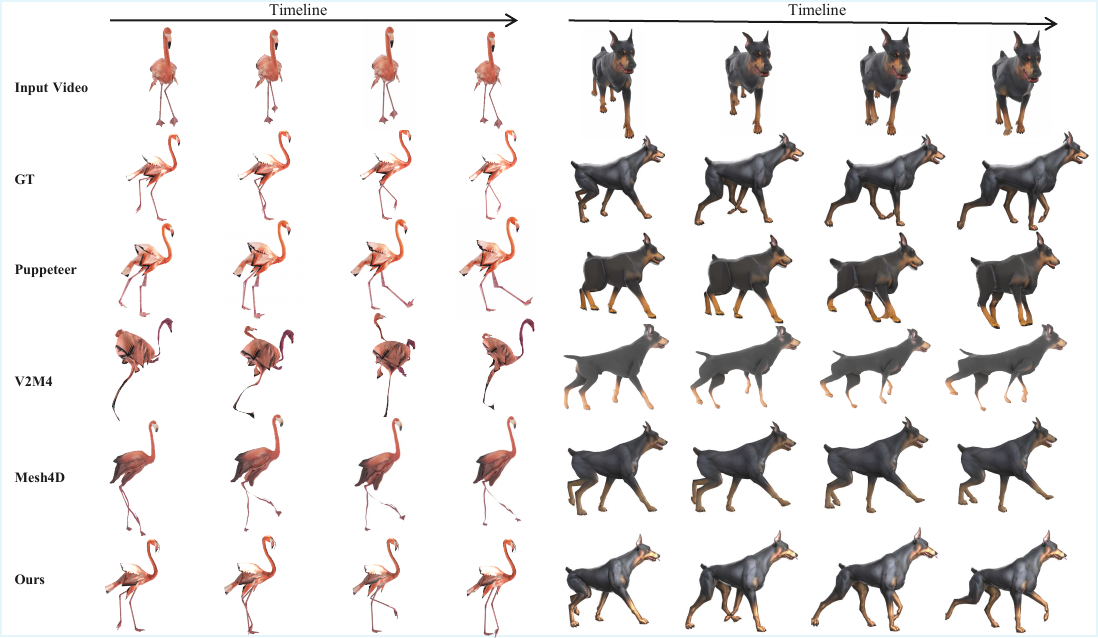}
  \caption{\textbf{Visilization results.} 
In our benchmark, we have access to ground-truth 4D assets, which allows us to render the input monocular video and the corresponding videos of viewpoints for evaluation.
We compare to baselines with different settings: \textbf{Puppeteer} is initialized with the provided GT 3D asset information; 
\textbf{V2M4} reconstructs a 3D object per frame and refine them; 
\textbf{Mesh4D} initializes a 3D asset from the first frame.
In contrast, \textbf{ours} starts from the same GT static asset extracting the rigged skeleton and skinning weights, and applies motion from extracted 3D trajectories, yielding more temporally consistent and structurally plausible animations.
  }
  \label{fig:qual_comp}
\end{figure}

4D generation extends 3D representations by adding a temporal dimension, where each frame exhibits distinct motion states of 3D objects. This advancement promises to significantly reduce animators' workload while catalyzing progress in AR, VR, and computer graphics applications.

Recent 4D generation methods typically start from two kinds of settings. Some methods synthesize 3D assets from text or image inputs and then animate them\cite{ct4d, 4dynamic, textmesh4d, dreamfusion, progressiverenderingdistillation, sam3, ar1to3}, offering a convenient way for content creation but often facing challenges in asset quality, structural consistency, or motion smoothness. Others begin with existing or reconstructed 3D assets and focus on motion transfer or animation under text or video guidance\cite{dreammesh4d, l4gm, ar4d, mesh4d, animate3d, bringing, driveanymesh, aam, puppeteer, camo, animamimicimitating3danimation}, which generally provides higher visual fidelity. Despite operating on different starting points, both converge on a common consense: first acquire a 3D asset, then prepare a structural or geometric representation for animation, and finally synthesize motion from text, video, or other control signals. This decomposition is attractive because it allows a model to inherit advances from mature text-to-3D and image-to-3D generators\cite{dreamfusion, progressiverenderingdistillation, sam3, ar1to3}, automatic rigging tools\cite{riganything, unirig, asmr, physrig}, and real or generated videos that include broad world knowledge about objects moving patterns\cite{stablevideodiffusionscaling, magicmotion}. 

However, two critical challenges emerge. The first is \textbf{representation}. Existing 4D animation often rely on dense meshes, Gaussian primitives, or other compact representations\cite{mesh4d, v2m4, aam}. Such representations are expensive to animate over long horizons, hard to obtain identification, and difficult to control in a structured manner. Therefore, generation is often limited to relatively short clips and easy to cause artifacts. The second is \textbf{motion prior}. Text prompts provide high-level semantic intent, but usually lack the fine-grained details of execution, such as timing, coordination, and speed variation. Video offers richer motion evidence, yet it entangles motion with appearance and background, making the transfer from observations to animation hard. 


These challenges call for rethinking both the representation and the motion prior: the former should be compact and temporally stable over long horizons, while the latter should be more explicit than text and more motion-centric than video. Inspired by these, we advocate to adopt skeleton as the representation and 3D point trajectories as the motion control signal for 4D generation. Skeletons provide a sparse and structured description and decouple appearance, making motion generation more compact, stable, and controllable.  Trajectories provides abundant geometric constraints and focus on motion patterns without appearance, enabling robust track-following even when individual tracks are noisy or sparse. Benefiting from recent advances in large-scale 3D point tracking models\cite{sptrack, cotracker3}, such trajectories can serve as a robust and informative motion prior. 

Based on these perspectives, we propose \texttt{ACT} (Skeleton \textbf{A}nimation \textbf{C}ontrolled by \textbf{T}rajectories), a trajectory-conditioned framework for topology-general skeletal animation. ACT receives as input 3D trajectories and a target skeleton derived from an existing static 3D asset. The trajectories are extracted from monocular video—either real-world footage or generated by video models—and serve as control signals for transferring motion patterns to the target skeleton via a Routed Trajectory Injector designed for both spatial and temporal alignment. For spatial alignment, we combine prior-guided hard routing and global routing: skinning-weight-guided routing establishes precise mappings between skeleton joints and mesh vertices which correlate with track points. Global routing enables bidirectional queries between all track points and skeleton joints for holistic pose awareness. For temporal alignment, we introduce local windowed cross-attention so that joints focus on nearby motion patterns, improving micro-timing and reducing misalignment across different motion rates. The fused features are then processed by topology-aware spatio-temporal attention modules to synthesize motion over time. To further support generalization across heterogeneous skeletons, ACT incorporates topology-agnostic normalization, including track-guided root normalization and offset-based normalization for non-root joints. Extensive experiments demonstrate that ACT outperforms existing methods.

In summary, our contributions are as follows:
\begin{itemize}
    \item \textbf{A trajectory-conditioned framework for topology-general skeletal animation.}
    We introduce ACT, which transfers motion from monocular video to skeletons of arbitrary topology through 3D trajectory control.
    
    \item \textbf{A Routed Trajectory Injector for robust animation.}
    We design a trajectory-to-joint injection module that combines hard routing for grounded correspondence, global routing for holistic context, and local temporal interaction for stable motion alignment.
    
    \item \textbf{Topology-agnostic normalization for cross-topology generalization.}
    We propose normalization strategies for root and non-root joints that improve skeleton animation without relying on dataset-level topology statistics.
\end{itemize}

\section{Related Work}

\subsection{Representation and 4D Generation}
Time-varying 3D content is commonly represented as explicit surfaces like meshes or Gaussian splatting, where the choice of representation largely determines the trade-off between rendering fidelity and computational cost.

\textit{Mesh-based} approaches remain appealing because they could output assets compatible with standard graphics pipelines. They can be broadly grouped into \textit{optimization-based} and \textit{feed-forward} paradigms. 
Optimization-based methods often yield high-quality, engine-friendly meshes but can be computationally heavy and sensitive to fast motion or occlusions, e.g., V2M4\cite{v2m4}. 
In contrast, feed-forward methods always enable one-shot  inference at the cost of relying on learned priors and potentially reduced robustness like include Mesh4D\cite{mesh4d} , AnimateAnyMesh\cite{aam}, DriveAnyMesh\cite{driveanymesh}, AnimaX\cite{animx}, and TriDiff-4D\cite{sheung2025tridiff}. They all face heavy presentation and computation costs which restricts mesh-based methods to produce long sequence results.

\textit{Gaussian-based} representations offer an alternative axis of trade-off, prioritizing high-quality rendering and flexible deformation modeling. Starting from 4D Gaussian Splatting\cite{4dgs}, many works learn time-varying deformation fields or dynamic attributes for Gaussians and optimize them under video diffusion supervision, often via Score Distillation Sampling (SDS), which is effective but can suffer from randomness-induced instability and temporal drift, like Consistent4D\cite{consistent4d}, SC4D\cite{sc4d}, MVTokenFlow\cite{mvtokenflow}, In-2-4D\cite{in24d}. Recent directions explore more structured generation and training strategies, such as SDS-free autoregressive 4D generation like AR4D\cite{ar4d} or mutual optimization between video and 4D like Video4DGen\cite{video4dgen}. 
Meanwhile, learning compact latent spaces for canonical Gaussians and their temporal variations enables feed-forward video-to-4D synthesis without per-instance fitting like Gaussian Variation Field Diffusion\cite{zhang2025gaussian}. Despite strong rendering quality, purely deformation-driven Gaussians typically lack explicit kinematic structure, making precise articulated control non-trivial.

This motivates \textit{hybrid} models that inject skeletal priors into 4D representations. Some works bind Gaussians together with skeletons for controllable animation including DRiVE\cite{drive} and Make-It-Animatable\cite{makeitanimatable}, others use pose sequences as explicit control signals for 4D character animation like CharacterShot\cite{charactershot}, or distill articulated motion into low-DoF skeleton parameters using video diffusion supervision as AKD\cite{AKD}. The above tells that skeleton remains useful tool to carry motion pattern.

\subsection{Controllable Motion Synthesis}
Control signals for motion synthesis range from sparse semantics to dense observations, providing important motion information.

\textit{Text-driven} control models offers semantic intent but is inherently underspecified for micro-timing and joint-level coordination like AnimateAnyMesh\cite{aam}, TriDiff-4D\cite{sheung2025tridiff}, GaussianMotion\cite{gaussianmotion}, X-MoGen \cite{xmogen}, TextMesh4D\cite{textmesh4d}. To strengthen structure, many pipelines introduce explicit pose or skeleton control like CharacterShot\cite{charactershot}, AKD\cite{AKD}, Make-It-Animatable\cite{AKD}. 

\textit{Video-driven} control provides temporally dense cues but entangles motion with appearance, lighting, and background—often requiring heavy optimization or strong priors to avoid drift like Consistent4D\cite{consistent4d}, SC4D\cite{sc4d}, MVTokenFlow\cite{mvtokenflow}, Video4DGen\cite{video4dgen}, DriveAnyMesh\cite{driveanymesh}. Systems that explicitly aim to output reusable animation assets often integrate rigging and optimization, such as Puppeteer\cite{puppeteer}, while category-agnostic motion capture approaches such as MoCapAnything\cite{mocapanything} recover joint trajectories and then solve asset-specific rotations via inverse kinematics, improving cross-topology applicability.

\textit{Track-based} control provides a middle ground: it is temporally dense yet structurally sparse and appearance-invariant. Prior work has shown that trajectories can substantially expand motion controllability in generative models like TC4D\cite{bahmani2024tc4d} and Motion Prompting\cite{motionprompt} and can support both global scene motion and object-level motion constraints. Nevertheless, existing trajectory conditioning is often applied as global transforms or unstructured points, rather than being systematically fused with articulated skeletal dynamics for arbitrary topologies. 

\section{Method}
\label{sec:method}

We present a topology-general skeletal motion generator ACT controlled by 3D tracks extracted from the monocular video. Given a target skeleton of arbitrary topology and a set of tracks, ACT predicts the skeleton motion sequence by (i) precomputing tracks spatiotemporal feature and joints feature, (ii) injecting track information into joint tokens through a Routed Trajectory Injector, and (iii) refining motion with topology-aware spatiotemporal modules. 

\subsection{Preliminaries}
\label{sec:preliminaries}

\paragraph{AnyTop backbone.}
We adopt AnyTop~\cite{anytop} as the topology-general backbone to represent and model skeletal motion. Specifically, the sequence is encoded as joint tokens $\mathbf{X}\in\mathbb{R}^{T\times J\times D}$ padded to $(T_{\max},J_{\max})$ with masks, where each joint at each frame forms an individual token. We follow AnyTop’s \textit{root-invariant} coordinate to remove global translation bias by expressing positions relative to the root:
$\hat{\mathbf{p}}_j^t=\mathbf{p}_j^t-\mathbf{p}_{\text{root}}^t$.
A skeleton is represented as a DAG
$\mathcal{S}=\{\mathbf{O},\mathbf{R}_{\mathcal{S}},\mathbf{D}_{\mathcal{S}},\mathbf{N}_{\mathcal{S}}\}$,
where $\mathbf{O}$ are rest offsets, $\mathbf{R}_{\mathcal{S}}$ encodes discrete joint relations, $\mathbf{D}_{\mathcal{S}}$ stores shortest-path graph distances, and $\mathbf{N}_{\mathcal{S}}$ contains joint names. AnyTop initializes joint tokens $\mathbf{H}^{(0)}$ by combining offsets information. Motion dynamics are then refined by topology-aware spatio-temporal transformer module: spatial attention is biased by relation and distance priors via learnable embeddings, and temporal attention enforces long-horizon coherence conditioned preserving joint identity and structural consistency across diverse skeleton topologies.

\begin{figure*}[t]
  \includegraphics[width=\textwidth]{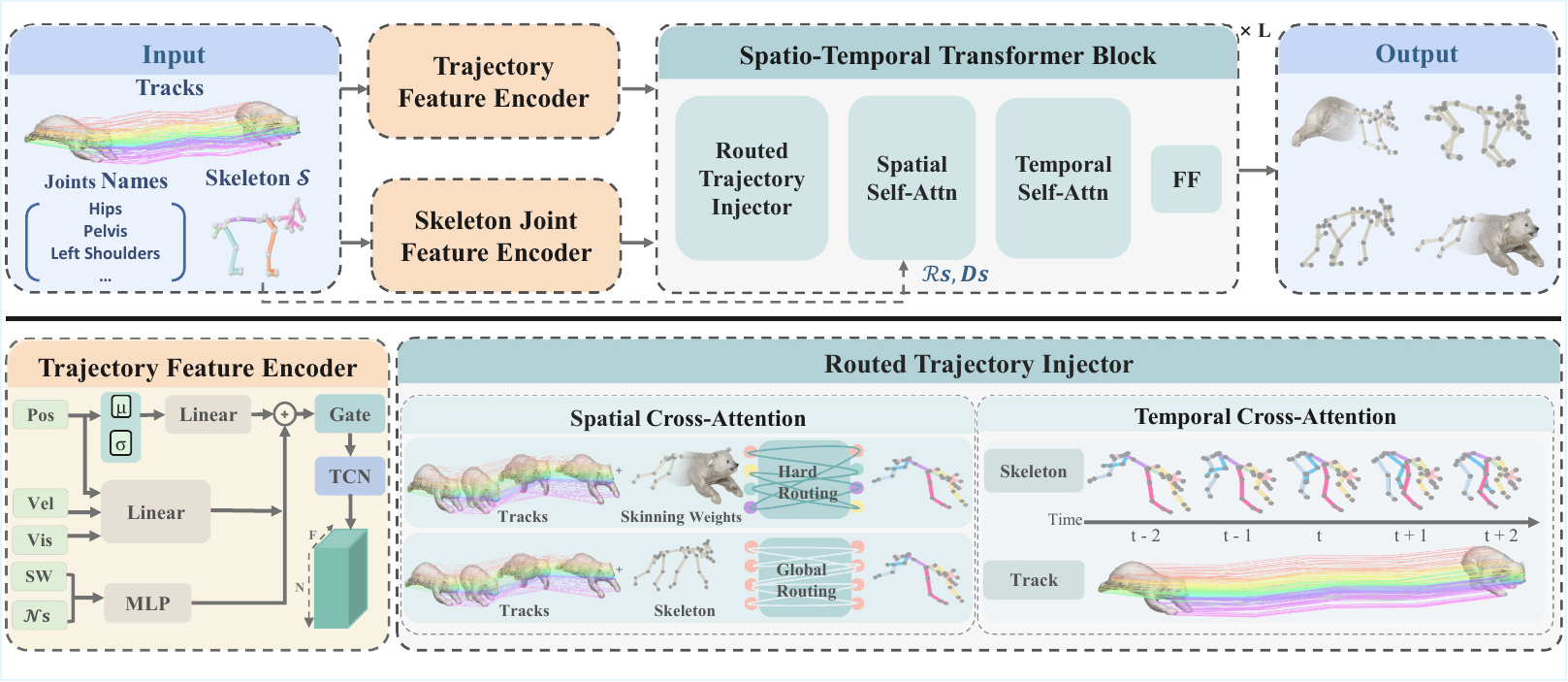}
  \caption{\textbf{Overview of ACT framework.} 
  ACT encodes 3D trajectories and arbitrary skeletons into track and joint tokens, respectively. 
  To effectively fuse these modalities, the \textbf{Routed Trajectory Injector (RTI)} employs a dual-path mechanism: 
  (1) \textbf{Hard Routing} leverages geometric skinning priors for local track-to-joint alignment; 
  (2) \textbf{Global Routing} captures holistic context via cross-attention. 
  A \textbf{Temporal Refinement} module is integrated to ensure temporal coherence. 
  Finally, topology-aware modules refine the motion sequence by enforcing structural constraints.}
  \label{fig:model}
\end{figure*}

\paragraph{Track representation.}
Given an input video, we extract a set of $K$ motion tracks and represent them as $\mathbf{P}\in\mathbb{R}^{T\times K\times 3}$.
We pad tracks to $K_{\max}$ and associate two masks: a static mask $\mathbf{m}_{\text{static}}\in\{0,1\}$ indicating invalid/valid tracks, and a dynamic probability $\mathbf{v}\in[0,1]$ as visibility probability.
When available, we additionally use a skinning-based prior that links tracks to the skeleton.
We denote the top-$K_s$ candidate joints and their normalized skinning weights by
$\mathbf{S}_{\text{idx}}\in\{-1,\dots,J-1\}^{K_{\max}\times K_s}$ and $\mathbf{S}_{\text{w}}\in\mathbb{R}^{K_{\max}\times K_s}$, respectively, where $-1$ indicates padding/invalid entries and $\sum_{s=1}^{K_s}\mathbf{S}_{\text{w}}(k,s)=1.$

\subsection{Architecture}
\label{sec:arch}

Our architecture consists of these components: a \emph{Skeleton Joint Feature Encoder} that encodes the skeleton into joint tokens, a \emph{Track Feature Extractor} that converts raw trajectories into spatiotemporal track tokens, a \emph{Routed Trajectory Injector (RTI)} that injects track motion information into joints, and \emph{Spatio-Temporal  Module} consists of two attentions that model structured dynamics over time and output final results.

\subsubsection{Normalization.}

A challenge in topology-general skeleton animation is to normalize skeleton features without injecting topology-specific biases. Unlike AnyTop~\cite{anytop}, which uses dataset-level statistics that may entangle topology priors we adopt a topology-agnostic normalization guided by tracks and intrinsic offsets.

\paragraph{Track-guided root normalization.}
To make root translation preserving the motion scale implied by the observed tracks, we normalize the root trajectory using sequence-level track statistics. Specifically, given all tracked 3D points $\{\mathbf{p}_k^t\}$ in a sequence, we compute their mean location and overall spatial scale as
\begin{equation}
\boldsymbol{\mu}_{\text{track}}=\frac{1}{KT}\sum_{k,t}\mathbf{p}_k^t,\quad
\boldsymbol{\sigma}_{\text{track}}=\sqrt{\frac{1}{KT}\sum_{k,t}\|\mathbf{p}_k^t-\boldsymbol{\mu}_{\text{track}}\|^2}.
\end{equation}
The root translation at frame $t$ is then normalized by
\begin{equation}
\tilde{\mathbf{p}}_{\text{root}}^{t}=\frac{\mathbf{p}_{\text{root}}^{t}-\boldsymbol{\mu}_{\text{track}}}{\boldsymbol{\sigma}_{\text{track}}+\epsilon}.
\end{equation}
Compared with normalization based only on joints statistics, this formulation ties the root motion to the motion of the input tracks, which better preserves motion magnitude and yields more stable guidance across object types.

\paragraph{Offset-based non-root normalization.}
For non-root joints in root-relative coordinates, we use skeleton-intrinsic offsets as a mean shift:
\begin{equation}
s_j=\mathbf{o}_j+\mathbf{o}_{\text{parent}(j)},\qquad
\tilde{\mathbf{p}}_{j}^{t}=\hat{\mathbf{p}}_{j}^{t}-s_j,\ \ j\neq \text{root}.
\end{equation}
We intentionally avoid variance normalization from bone lengths to prevent instability for extremely short bones. Overall, this normalization maintains consistent scale across skeletons without relying on dataset-wide statistics.

\subsubsection{Track Feature Encoder}
\label{sec:trackpre}

Given tracks $\mathbf{P}\in\mathbb{R}^{B\times K_{\max}\times T\times 3}$, we compute per-track spatiotemporal tokens. 

We first encode three complementary streams: position, velocity, and visibility. We compute per-frame velocity by finite difference with zero-padding,
$\mathbf{V}_{b,n,t}=\mathbf{P}_{b,n,t}-\mathbf{P}_{b,n,t-1}$,
and project each stream into the model dimension $D$:
\begin{equation}
\mathbf{F}_{\text{pos}}=\mathbf{P}\mathbf{W}_{\text{pos}},\quad
\mathbf{F}_{\text{vel}}=\mathbf{V}\mathbf{W}_{\text{vel}},\quad
\mathbf{F}_{\text{vis}}=\mathbf{v}\mathbf{W}_{\text{vis}},
\end{equation}
where $\mathbf{W}_{(\cdot)}$ are learned linear projections. The three streams are fused and stabilized by layer normalization and dropout:
\begin{equation}
\mathbf{F}=\mathrm{Drop}\!\left(\mathrm{LN}\left(
\mathbf{F}_{\text{pos}}+\mathbf{F}_{\text{vel}}+\mathbf{F}_{\text{vis}}
\right)\right)\in\mathbb{R}^{B\times K_{\max}\times T\times D}.
\end{equation}

When skinning-based priors are available, We use the top-$K_s$ skinning indices/weights $(\mathbf{S}_{\text{idx}},\mathbf{S}_{\text{w}})$ and compress them with a small MLP:
\begin{equation}
\bar{\mathbf{s}}_{b,n}=\sum_{k=1}^{K_s}\mathbf{S}_{\text{w}}(b,n,k)\,
\mathbf{S}_{b,\mathbf{S}_{\text{idx}}(b,n,k)},\qquad
\mathbf{f}^{\text{sw}}_{b,n}=\phi_{\text{sw}}(\bar{\mathbf{s}}_{b,n})\in\mathbb{R}^{D}.
\end{equation}
We then broadcast-add $\mathbf{f}^{\text{sw}}_{b,n}$ to $\mathbf{F}_{b,n,:,:}$ across time, injecting a lightweight semantic prior.

Track sets extracted from videos are noisy and uneven: some tracks are short or occluded. Instead of discarding them with hard thresholds, we learn a soft importance gate $a_{b,n}$ to down-weight uninformative tracks. Using the validity mask $\mathbf{m}_{b,n,t}$, we compute masked statistics and predict the gate:
\begin{equation}
\boldsymbol{\mu}_{b,n}=\frac{\sum_t \mathbf{m}_{b,n,t}\mathbf{P}_{b,n,t}}{\sum_t \mathbf{m}_{b,n,t}+\epsilon},\quad
\boldsymbol{\sigma}_{b,n}=\sqrt{\frac{\sum_t \mathbf{m}_{b,n,t}\left(\mathbf{P}_{b,n,t}-\boldsymbol{\mu}_{b,n}\right)^2}{\sum_t \mathbf{m}_{b,n,t}+\epsilon}},
\end{equation}
\begin{equation}
    \quad
a_{b,n}=\psi([\boldsymbol{\mu}_{b,n},\boldsymbol{\sigma}_{b,n}]),
\end{equation}
and rescale the whole sequence as $\mathbf{F}_{b,n,:,:}\leftarrow a_{b,n}\mathbf{F}_{b,n,:,:}$, which makes subsequent routing and attention more robust to distractor tracks.

Finally, since frame-wise features are still locally embedded, we apply a lightweight non-causal temporal convolution (TCN) along the time axis of each track to aggregate short-range context and smooth micro-timing cues:
\begin{equation}
\mathbf{T}^{(0)}=\mathrm{TCN}(\mathbf{F})\in\mathbb{R}^{B\times K_{\max}\times T\times D}.
\end{equation}
$\mathbf{T}^{(0)}$ serves as the precomputed track context passed to all subsequent transformer blocks.

\subsubsection{Routed Trajectory Injector (RTI)}
\label{sec:rti}

Given intermediate motion tokens $\mathbf{H}^{(0)}\in\mathbb{R}^{T\times B\times J_{\max}\times D}$ and precomputed track context $\mathbf{T}^{(0)}\in\mathbb{R}^{B\times K_{\max}\times T\times D}$, RTI injects track information into joints by combining complementary spatial alignment paths and a temporal refinement stage. The design goal is: (i) resolve the intrinsic ambiguity of mapping a set of tracks to a sparse set of joints, and (ii) distill a stable motion signal under tracks through time window.

\paragraph{Spatial alignment.}
RTI aligns track context to joints at each frame using two complementary paths: (i) prior-guided hard routing based on skinning correspondences, and (ii) global routing via cross-attention for holistic pose awareness.
We denote joint tokens as $\mathbf{H}^{(0)}\in\mathbb{R}^{T\times B\times J_{\max}\times D}$ and track context as $\mathbf{T}^{(0)}\in\mathbb{R}^{B\times K_{\max}\times T\times D}$.
For convenience, we use per-frame slices
$\mathbf{H}^{t,b}=\mathbf{H}^{(0)}_{t,b,:,:}\in\mathbb{R}^{J_{\max}\times D}$ and
$\mathbf{T}^{t,b}=\mathbf{T}^{(0)}_{b,:,t,:}\in\mathbb{R}^{K_{\max}\times D}$.

\textbf{Hard routing.}
We form a track-to-joint assignment matrix $\mathbf{A}\in\mathbb{R}^{B\times K_{\max}\times J_{\max}}$ by scattering the sanitized skinning indices and weights:
\begin{equation}
\mathbf{A}_{b,n,j} =
\frac{\sum_{k=1}^{K_s}\mathbb{I}\!\big(\mathbf{S}_{\text{idx}}(b,n,k)=j\big)\,\mathbf{S}_{\text{w}}(b,n,k)}
{\sum_{j'}\sum_{k=1}^{K_s}\mathbb{I}\!\big(\mathbf{S}_{\text{idx}}(b,n,k)=j'\big)\,\mathbf{S}_{\text{w}}(b,n,k)+\epsilon}.
\end{equation}
The routed joint-aligned feature is computed by aggregating tracks into each joint at every frame:
\begin{equation}
\mathbf{C}^{t,b}_{\text{route}}(j,:)=\sum_{n=1}^{K_{\max}}\mathbf{A}_{b,n,j}\,\mathbf{T}^{(0)}_{b,n,t,:},
\qquad
\mathbf{C}_{\text{route}}\in\mathbb{R}^{T\times B\times J_{\max}\times D}.
\end{equation}
In practice, invalid/static tracks are masked out when constructing $\mathbf{A}$ or $\mathbf{T}^{(0)}$.

\textbf{Global routing.}
We additionally attend from joints to all tracks at each frame:
\begin{equation}
\mathbf{C}^{t,b}_{\text{glob}}
=
\mathrm{MHA}\!\left(\mathbf{Q}=\mathbf{H}^{t,b},\,\mathbf{K}=\mathbf{T}^{t,b},\,\mathbf{V}=\mathbf{T}^{t,b}\right)
\in\mathbb{R}^{J_{\max}\times D},
\end{equation}
where key padding masks are derived from $\mathbf{m}_{\text{static}}$ and track validity to prevent attending to invalid or static tracks.

\textbf{Fusion.}
We fuse the two aligned features with learnable gates and inject them residually:
\begin{equation}
\mathbf{C}^{t,b}=\gamma_r\,\mathbf{C}^{t,b}_{\text{route}}+\gamma_G\,\mathbf{C}^{t,b}_{\text{glob}},
\qquad
\mathbf{H}'^{t,b}=\mathrm{LN}\!\left(\mathbf{H}^{t,b}+\mathbf{C}^{t,b}\right),
\end{equation}
yielding $\mathbf{H}'\in\mathbb{R}^{T\times B\times J_{\max}\times D}$ for subsequent temporal refinement.

\paragraph{Temporal alignment.}
Spatial injection alone is frame-local and may miss temporal information such as rhythm, micro-timing, and velocity, especially when tracks are missing under occlusion. We therefore add a temporal refinement stage that lets each joint token selectively attend to its injected motion track features over a local time window. Concretely, we reshape the joint stream from $\mathbf{H}'\in\mathbb{R}^{T\times B\times J_{\max}\times D}$ into $
\mathbf{Q}_{\text{temp}}=\mathrm{reshape}(\mathbf{H}')\in\mathbb{R}^{(B J_{\max})\times T\times D}$,
so that each joint becomes an independent temporal sequence in the batch dimension, making temporal modeling explicit and efficient. We use the fused injected feature sequence as keys/values,
$\mathbf{K}_{\text{temp}}=\mathbf{V}_{\text{temp}}=\mathrm{reshape}(\mathbf{C})\in\mathbb{R}^{(B J_{\max})\times T\times D}$.
We add aligned positional encodings to preserve temporal order. Temporal cross-attention is then applied within a local window mask $\mathbf{M}_{\text{win}}$ to focus on short-range dynamics and to reduce sensitivity to long-range missing segments:
\begin{equation}
\Delta\mathbf{H}_{\text{temp}}
=
\mathrm{MHA}\!\left(\mathbf{Q}_{\text{temp}},\mathbf{K}_{\text{temp}},\mathbf{V}_{\text{temp}};\ \mathbf{M}_{\text{win}}\right),
\end{equation}
\begin{equation}
\mathbf{H}''_{\text{temp}}=\mathrm{LN}\!\left(\mathbf{Q}_{\text{temp}}+\gamma_t\,\Delta\mathbf{H}_{\text{temp}}\right),
\end{equation}
where $\gamma_t$ is a learned gate and $\mathbf{M}_{\text{win}}$ restricts attention to a radius-$r$ temporal neighborhood. Finally, we reshape $\mathbf{H}''_{\text{temp}}$ back to $\mathbf{H}''\in\mathbb{R}^{T\times B\times J_{\max}\times D}$. This temporal refinement corrects residual misalignment from the spatial stage and encourages temporally coherent joint dynamics even under imperfect tracks.

\subsection{Training Objectives}
\label{sec:training}

Given a motion sequence $\mathbf{X}_0$ of skeleton $S$, with a diffusion step $\tau \sim [1, T_d]$,
our model predicts the clean motion $\hat{\mathbf{X}}_0 = \text{ACT}(\mathbf{X}_\tau, \tau, S)$.
\begin{equation}
\mathcal{L}_{\text{simple}}=\mathbb{E}_{\tau \sim [1,T_d]}
\left\|\text{ACT}(\mathbf{X}_\tau,\tau,S)-\mathbf{X}_0\right\|_2^2.
\end{equation}

To improve the modeling of rotations, we observe that the MSE over continuous 6D representations does not strictly reflect the true angular distance in the rotation space. Therefore, we project the network outputs back to the unnormalized physical domain and apply a stabilized geodesic loss over the rotation matrices. Let $\mathbf{R}$ and $\hat{\mathbf{R}} \in \mathbb{R}^{N \times J \times 3 \times 3}$ represent the rotation matrices derived from the ground truth and predicted 6D representations using the Gram-Schmidt process \cite{rotation}. The standard geodesic distance involves an $\arccos$ operation, which suffers from numerical instability and exploding gradients when $\hat{\mathbf{R}} \to \mathbf{R}$. To ensure stable training, we formulate our rotation loss based on the cosine of the angular difference:
\begin{equation}
\mathcal{L}_{\text{rot}} = \sum_{n=1}^{N} \sum_{j=1}^{J} \omega_j \left( 1 - \frac{\text{Tr}(\mathbf{R}_{n,j} \hat{\mathbf{R}}_{n,j}^T) - 1}{2} \right),
\end{equation}
where $\text{Tr}$ is the matrix trace operation, and $\omega_j$ assigns different weights to the root joint and non-root joints to prioritize the global orientation accuracy. This smooth surrogate function avoids the singularity of $\arccos$.

Finally, the complete training objective is a weighted sum of the simple loss and the geodesic loss:
\begin{equation}
\mathcal{L} = \mathcal{L}_{\text{simple}} + \lambda_{\text{rot}} \mathcal{L}_{\text{rot}},
\end{equation}
where $\lambda_{\text{rot}}$ is a hyperparameter that controls the relative importance of the rotation loss in the final objective. This approach ensures that both the clean motion prediction and the rotational accuracy are optimized during training.

\section{Experiments}
\label{sec:experiments}

\subsection{Datasets and Implementation Details}
\label{sec:datasets}

\begin{figure}[t]
  \centering
  \includegraphics[width=\columnwidth]{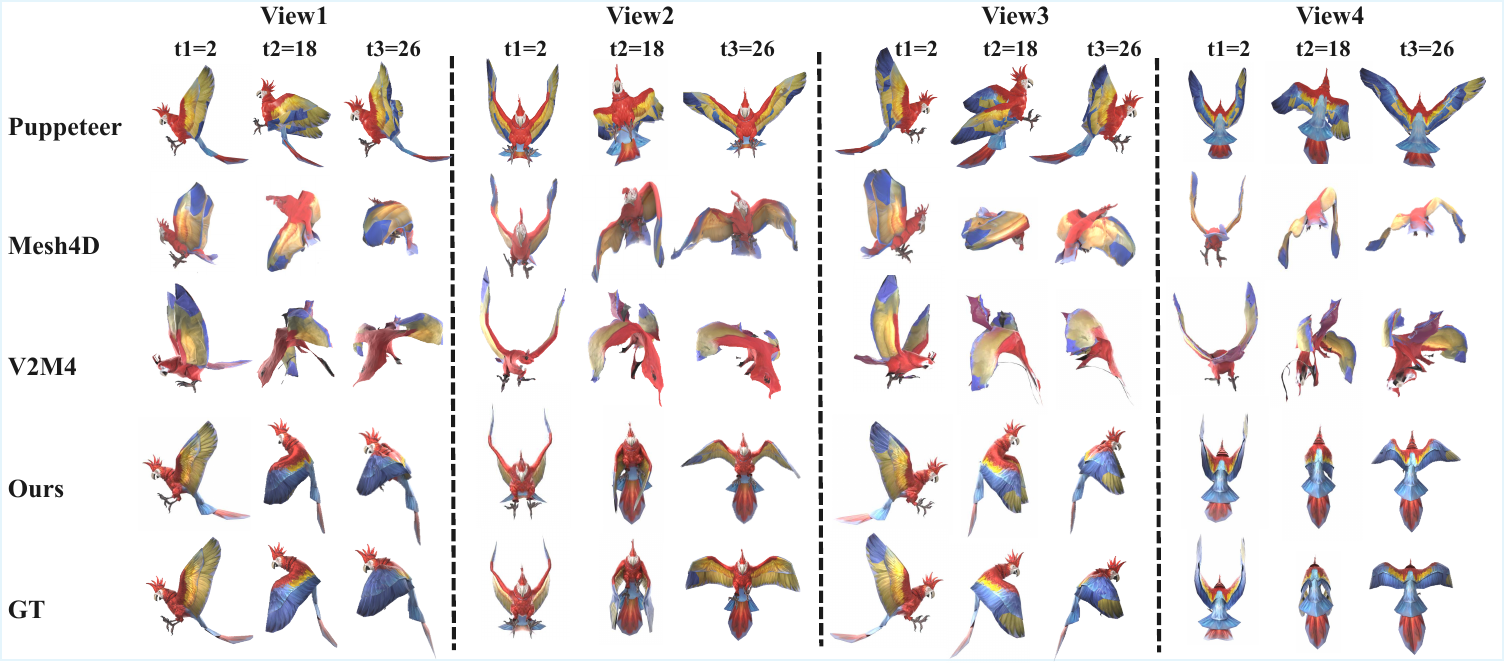}
  \caption{\textbf{Multiple views of qualitative comparison.} 
  Compare our ACT with Puppeteer, Mesh4D, and V2M4 across multiple viewpoints and time steps. 
  }
  \label{fig:qual_multiview}
\end{figure}

We evaluate our model on two distinct motion datasets: Truebones for diverse animal motion and AMASS\cite{amass} for large‑scale human motion.
The Truebones dataset contains a wide variety of high‑quality animated animal motion clips, covering 70+ species and multiple categories of motion. In our experiments, we follow almost the same data pre-processing pipeline as described in the AnyTop~\cite{anytop}. For evaluation, we select a set of complete and representative animal samples from each category to test the model’s ability to generate realistic and temporally consistent motion among unseen skeleton topologies.
To validate the generalization capabilities of our approach beyond non‑human motion, we also conduct experiments on the AMASS dataset. 
For human data, we tried two training ways to observe qualitative results, including sampling human data and mix it with Truebones data and training fully on human data.
We use transformer blocks L = 4 and $\tau$ = 100 diffusion steps. We train on 8 NVIDIA A100 GPUs.
More implementation details can be found in supplementary materials.

\begin{table}[t]
\centering
\scriptsize
\setlength{\tabcolsep}{4.5pt}
\renewcommand{\arraystretch}{1.1}
\begin{tabular}{l c S[table-format=1.3] S[table-format=1.3] S[table-format=1.4] S[table-format=1.4] S[table-format=4.1] S[table-format=2.2]}
\toprule
\textbf{Method} & \textbf{Sk.} &
{\textbf{Consist.}$\uparrow$} & {\textbf{Dy. Deg.}$\uparrow$} & {\textbf{M. Sm.}$\uparrow$} &
{\textbf{LPIPS}$\downarrow$} & {\textbf{FVD}$\downarrow$} & {\textbf{MPJPE}$\downarrow$} \\
\midrule
V2M4      & No & 0.886 & 0.610 & 0.9899 & 0.0910 & 806.8 & {\text{--}} \\
Puppeteer & Yes & 0.956 & 0.420 & 0.9908 & 0.1260 & 761.2 & 15.06 \\
Mesh4D    & No & 0.902 & 0.601 & 0.9908 & 0.1230 & 795.4 & {\text{--}} \\
\textbf{Ours} & Yes & \textbf{0.961} & \textbf{0.624} & \textbf{0.9967} & \textbf{0.0906} & \textbf{610.1} & \textbf{6.08} \\
\bottomrule
\end{tabular}
\caption{Quantitative comparison on the Truebones dataset. \textbf{Sk.}: whether the method uses an explicit skeleton. \textbf{Consist.}: subject consistency ($\uparrow$). \textbf{Dy. Deg.}: dynamic degree ($\uparrow$). \textbf{M. Sm.}: motion smoothness ($\uparrow$). \textbf{MPJPE}: mean per joint position error ($\downarrow$).}
\label{tab:metric_comp}
\end{table}

\subsection{Evaluation Metrics}
\label{sec:metrics}

We evaluate the rendered videos using three temporal-quality metrics as \emph{subject consistency}, \emph{motion smoothness}, and \emph{dynamic degree}\cite{huang2023vbench, zheng2025vbench2, huang2025vbench++}, together with two widely-used metrics, \emph{LPIPS}~\cite{lpips} and \emph{FVD}~\cite{fvd}. Unless otherwise stated, all metrics are computed on videos rendered with the same resolution, frame rate, and clip length. 
\textbf{Subject consistency} focuses on whether the appearance of the main subject remains stable throughout the video.
\textbf{Motion smoothness} focuses on whether the movement is temporally smooth and physically plausible, instead of only measuring the consistency of the look. 
\textbf{Dynamic degree} measures how much meaningful motion is present in a video to avoid overly-static outputs scoring high on consistency-related metrics. 
\textbf{LPIPS } measures perceptual distance between two images using deep features. We compute LPIPS on rendered frames and average over time for each video. 
\textbf{FVD} measures the distribution distance between generated and reference videos in a spatiotemporal feature space. Since FVD is sensitive to the number of frames, we report FVD under a fixed clip length for all methods. 
\textbf{MPJPE} measures the mean Euclidean distance between predicted and ground-truth joint positions, providing a direct assessment of joint-level motion accuracy. Since ground-truth skeleton annotations are only available for a subset of methods, we report MPJPE where applicable.

\begin{figure}[t] 
  \centering
  \includegraphics[width=\columnwidth]{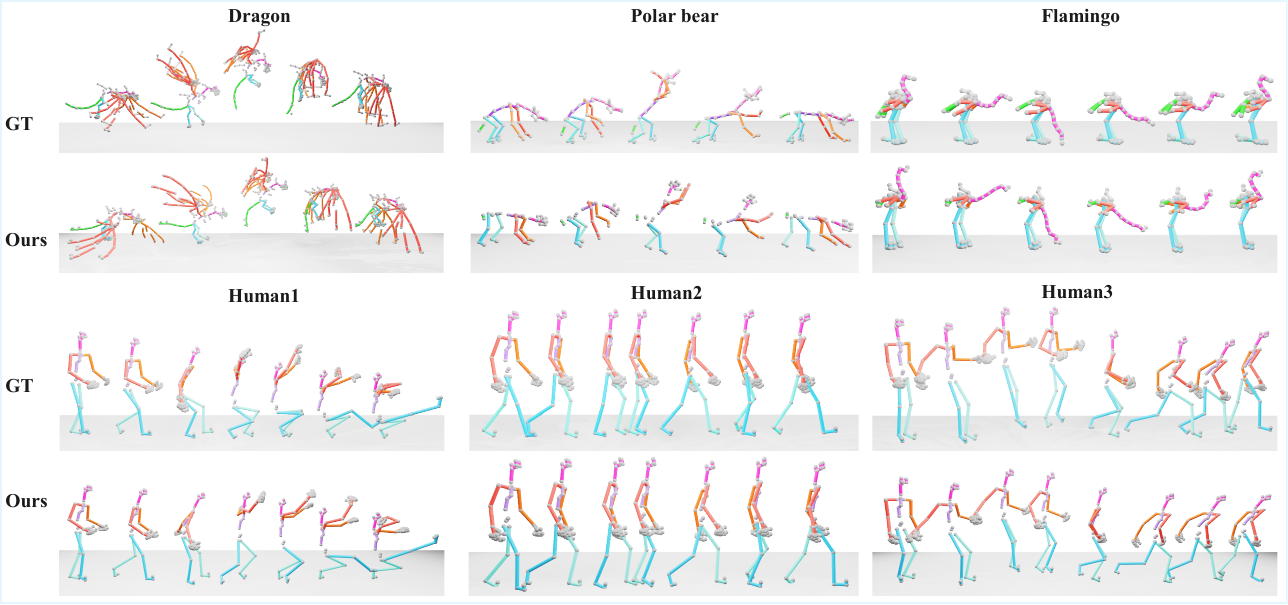}
  \caption{\textbf{Qualitative results of skeleton animation acrose categories.} 
  ACT successfully generates motions while maintaining temporal fidelity and joint structure.}
  \label{fig:qual_skeleton}
\end{figure}

\subsection{Baselines and Comparisons}
\label{sec:comparisons}
\paragraph{Baselines.}
We compare against representative mesh-based methods, including 
V2M4~\cite{v2m4}, Puppeteer~\cite{puppeteer}, and Mesh4D~\cite{mesh4d}.
These methods primarily produce time-varying geometry or rendered video results and typically do not output an explicit articulated skeleton motion that is directly comparable to track-driven skeletal animation.
Therefore, we conduct comparisons mainly in the rendered-video space: we render outputs under a unified setup and evaluate them using subject consistency, motion smoothness, and dynamic degree, together with LPIPS and FVD when paired references and protocols permit.
We also consider video-driven related work such as Drive Any Mesh~\cite{driveanymesh} and AnimaX~\cite{animx}, and skeleton animation work like MoCapAnything\cite{mocapanything}. However, since their official implementations are not publicly available at the time of writing, we do not include them in our comparison.

\paragraph{Comparison protocol and main results.}
For fair comparison, we try our best to keep the rendering protocol consistent across all methods, including distance, background, resolution, frame rate, and clip length. A practical mismatch exists for Mesh4D, which is generated at a sparse set of keyframes (e.g., 6 frames) and then temporally interpolated to match the evaluation length (e.g., 32 frames); we follow this standard pipeline and report results accordingly.
As summarized in Table~\ref{tab:metric_comp}, our method achieves the best performance on subject consistency (0.961), dynamic degree (0.624), motion smoothness (0.9967), LPIPS (0.0906), FVD (610.1) and MPJPE(6.08), outperforming all baselines on these metrics. Overall, our skeleton-driven representation yields a consistently better trade-off across temporal coherence, motion fidelity, and perceptual quality.

\begin{figure}[t]
  \centering
  \includegraphics[width=\linewidth]{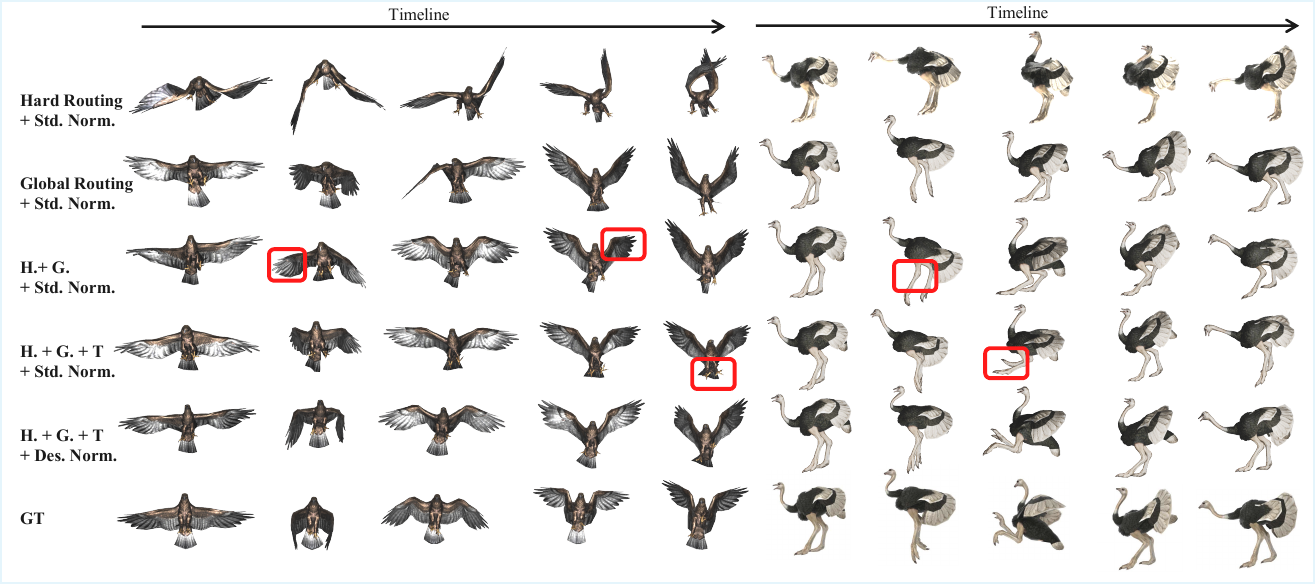}
  \caption{\textbf{Qualitative ablation results} evaluated in rendered-video space. 
  }
  \label{fig:ablation_qual}
\end{figure}

\paragraph{Qualitative analysis.}
 As shown in Fig.~\ref{fig:qual_comp}, Puppeteer produces incorrect articulation patterns due to the absence of skeletal priors, V2M4 exhibits limb flickering, and Mesh4D suffers from geometry collapsing between keyframes. In contrast, our ACT framework preserves consistent motion even under occlusion, owing to the explicit skeletal representation that enforces topological consistency across time.
Fig.~\ref{fig:qual_multiview} further validates temporal and geometric consistency across multiple viewpoints and time steps. While baselines exhibit distorted wings or unstable appendages under view changes, our method maintains structural coherence across all rendered views---a direct consequence of the globally-consistent track-driven skeleton that anchors deformation across both time and space.
Finally, Fig.~\ref{fig:qual_skeleton} demonstrates skeleton motion generation quality across categories as dragon, polar bear, flamingo, and human, producing plausible joint configurations and capturing category-specific motion dynamics.

 PJPE also decreases steadily from 12.43 ($\mathbf{H.}$ only) to 8.15 ($\mathbf{H.}+\mathbf{G.}$), confirming improved joint-level accuracy.\subsection{Ablations}
\label{sec:ablation}

\begin{table}[t]
\centering
\scriptsize
\setlength{\tabcolsep}{4pt}
\renewcommand{\arraystretch}{1.1}
\begin{tabular}{
l
S[table-format=1.3]
S[table-format=1.2]
S[table-format=1.4]
S[table-format=1.3]
S[table-format=3.2]
S[table-format=2.2]
}
\toprule
\textbf{Variant} &
{\textbf{Consist.}$\uparrow$} &
{\textbf{Dy. Deg.}$\uparrow$} &
{\textbf{M. Sm.}$\uparrow$} &
{\textbf{LPIPS}$\downarrow$} &
{\textbf{FVD}$\downarrow$} &
{\textbf{MPJPE}$\downarrow$} \\
\midrule
Hard ($\mathbf{H.}$) + Std. Norm.   & 0.915 & 0.41 & 0.9951 & 0.120 & 850.20 & 12.43 \\
Global ($\mathbf{G.}$) + Std. Norm. & 0.936 & 0.49 & 0.9961 & 0.100 & 800.80 & 9.78 \\
$\mathbf{H.}+\mathbf{G.}$ + Std. Norm. & 0.937 & 0.56 & 0.9964 & 0.099 & 701.28 & 8.15 \\
$\mathbf{H.}+\mathbf{G.}+\mathbf{T}$ + Std. Norm. & 0.959 & 0.58 & 0.9966 & 0.095 & 630.44 & 6.59 \\
\textbf{$\mathbf{H.}+\mathbf{G.}+\mathbf{T}$ + Des. Norm.} & \textbf{0.961} & \textbf{0.62} & \textbf{0.9969} & \textbf{0.090} & \textbf{610.10} & \textbf{6.08} \\
\bottomrule
\end{tabular}
\caption{\textbf{Quantitative ablation results} evaluated in rendered-video space. 
$\mathbf{H}$: prior-guided hard routing. $\mathbf{G}$: global routing. $\mathbf{T}$: temporal refinement. \textbf{MPJPE}: mean per joint position error ($\downarrow$).}
\label{tab:ablation_settings}
\end{table}

We perform controlled ablations to assess the contribution of each component in our ACT. All variants share the same basic settings.

\paragraph{Routing design.}
RTI injects track information into joint tokens via \textbf{prior-guided hard routing} ($\mathbf{H.}$) and \textbf{global routing} ($\mathbf{G.}$).
Using only $\mathbf{H.}$ relies heavily on the correspondence prior and may under-utilize informative tracks, resulting in weaker motion dynamics and inferior perceptual quality (LPIPS/FVD).
Replacing it with $\mathbf{G.}$ consistently improves fidelity (e.g., LPIPS $0.120\!\rightarrow\!0.100$, FVD $850.2\!\rightarrow\!800.8$) by allowing joints to attend to all tracks, but it can be more sensitive to distractor/noisy trajectories.
Combining them ($\mathbf{H.}+\mathbf{G.}$) yields the better controllability--robustness trade-off under the same normalization, notably increasing dynamic degree ($0.49\!\rightarrow\!0.56$) and reducing FVD ($800.8\!\rightarrow\!701.3$). 
MPJPE also decreases steadily from 12.43 ($\mathbf{H.}$ only) to 8.15 ($\mathbf{H.}+\mathbf{G.}$), confirming improved joint-level accuracy.

\paragraph{Temporal refinement.}
We then add $\mathbf{T}$, a windowed temporal cross-attention module that refines the injected motion signal to better capture micro-timing and suppress frame-wise jitter.
Compared to $\mathbf{H}+\mathbf{G}$, adding $\mathbf{T}$ improves all reported metrics, indicating more coherent transitions and temporally stable articulation.
MPJPE further drops from 8.15 to 6.59, reflecting more precise per-joint alignment.

\paragraph{Topology-agnostic normalization.}
Finally, replacing standard normalization with our designed normalization (Des. Norm.) further improves both fidelity and temporal consistency.
With the same $\mathbf{H}+\mathbf{G}+\mathbf{T}$ injector, Des. Norm. improves subject consistency ($0.959\!\rightarrow\!0.961$) and dynamic degree ($0.58\!\rightarrow\!0.62$), while achieving the lowest LPIPS/FVD ($0.095\!\rightarrow\!0.090$, $630.44\!\rightarrow\!610.10$).
This confirms that sequence-local, topology-agnostic scaling better aligns track-driven motion magnitude across heterogeneous skeletons and the lowest MPJPE (6.08), demonstrating that topology-agnostic normalization benefits joint-level accuracy across heterogeneous skeletons.

\paragraph{Qualitative visualization.}
For the qualitative results in Fig.~\ref{fig:ablation_qual}, we render the animated mesh driven by the predicted skeleton under the same rigging and camera settings. The red boxes mark representative failure cases (e.g., local wing/limb deformation or unstable transitions) that appear more frequently without complementary routing and are further reduced by temporal refinement, supporting the step-by-step improvements observed in Table~\ref{tab:ablation_settings}.

\section{Limitations}
Our approach still has several limitations. (1) Motion quality depends on the quality and coverage of input tracks; severe noise or poor spatial distribution can reduce motion quality. (2) There remains domain dependency: models trained mainly on animals or humans may require adaptation to transfer to new motion domains. (3) Although robust aggregation tolerates moderate noise, the current system still relies on relatively dense track sets for redundant control and stability, which increases cost.

\section{Conclusion}
We presented \texttt{ACT}, a trajectory-conditioned framework for topology-general skeletal animation. By adopting skeletons as a compact structured representation and 3D point trajectories as explicit motion guidance, ACT enables controllable long-horizon animation while avoiding the artifacts and computational costs of dense representations. The Routed Trajectory Injector robustly transfers trajectory cues to skeleton joints through complementary spatial and temporal alignment designs, and topology-agnostic normalization further enables generalization across heterogeneous skeleton structures. Experiments demonstrate consistent gains in motion fidelity and temporal consistency over baselines, with ablations verifying each component's contribution. We hope ACT encourages further exploration of trajectory-driven skeletal animation as a scalable interface for controllable 4D generation.

%
%
\bibliographystyle{splncs04}
\bibliography{main}
\end{document}